\documentclass[runningheads]{llncs}
\usepackage[T1]{fontenc}
\usepackage{graphicx}
\usepackage[hyphens]{url}
\usepackage{hyperref}

\usepackage{color}

\usepackage[textsize=scriptsize]{todonotes}

\usepackage{subfiles}

\usepackage[mode=buildnew]{standalone}

\begin{document}
%
\title{From Production Logistics to Smart Manufacturing: The Vision for a New RoboCup Industrial League}
%
\titlerunning{RoboCup Smart Manufacturing}

\author{
Supun~Dissanayaka\inst{1}\thanks{Authors are listed in alphabetical order.}
\and Alexander~Ferrein\inst{2}
\and Till~Hofmann\inst{3}
\and Kosuke~Nakajima\inst{4}
\and Mario~Sanz-Lopez\inst{5}
\and Jesus~Savage\inst{6}
\and Daniel~Swoboda\inst{3}
\and Matteo~Tschesche\inst{2}
\and Wataru~Uemura\inst{4}
\and Tarik~Viehmann\inst{3}
\and Shohei~Yasuda\inst{4}
}
\authorrunning{S.~Dissanayaka et al.}
%
\institute{
 King Mongkut's Institute of Technology Ladkrabang Thailand \\
\email{supun.di@kmitl.ac.th}
\and
FH Aachen University of Applied Sciences\\
\email{\{ferrein,tschesche\}@fh-aachen.de}\\
 \and
RWTH Aachen University \\
\email{\{till.hofmann,daniel.swoboda\}@ml.rwth-aachen.de}\\
\email{viehmann@kbsg.rwth-aachen.de}\\
\and
 Ryukoku University\\
 \email{\{kosuke, shohei\}@vega.elec.ryukoku.ac.jp,wataru@rins.ryukoku.ac.jp}
\and 
CNRS, University of Lille\\
\email{mario.sanz-lopez@univ-lille.fr}\\
\and
National Autonomous University of Mexico\\
\email{robotssavage@gmail.com}
}
\maketitle              

\begin{abstract}
The RoboCup Logistics League is a RoboCup competition in a smart factory scenario that has focused on task planning, job scheduling, and multi-agent coordination.
The focus on production logistics allowed teams to develop highly competitive strategies, but also meant that some recent developments in the context of smart manufacturing are not reflected in the competition, weakening its relevance over the years.
In this paper, we describe the vision for the \emph{RoboCup Smart Manufacturing League},
a new competition designed as a larger smart manufacturing scenario, reflecting all the major aspects of a modern factory.
It will consist of several tracks that are initially independent but gradually combined into one smart manufacturing scenario.
The new tracks will cover industrial robotics challenges such as assembly, human-robot collaboration, and humanoid robotics, but also retain a focus on production logistics.
We expect the reenvisioned competition to be more attractive to newcomers and well-tried teams, while also shifting the focus to current and future challenges of industrial robotics.

\end{abstract}

\section{Introduction}
The RoboCup Logistics League (RCLL) models a smart manufacturing environment,
in which participating teams deploy a fleet of mobile and autonomous robots to manufacture varying
products against the constraints of time, oversubscription, and uncertain
execution behaviour.
The key to success in the RCLL is long term autonomous behaviour 
without any human intervention as well as reliability and error-handling. 
%
The core challenge of the RCLL — its online planning and scheduling under real-world constraints and uncertain execution behavior — remains highly relevant.
At the same time, both research and industry are evolving and there has been rapid progress in the development of standardized robot platforms, factory automation, as well as paradigm shifts in terms of robot design, both on a hardware and software level.

In this paper, we propose the \emph{RoboCup Smart Manufacturing League (SML)}, a ground-up re-development of the RCLL to ensure its 
long-term viability as a robotics research platform and as a competition under the RoboCup umbrella.
To achieve this, we address two major points:
increasing accessibility of the league and supporting a broader spectrum of 
industry-relevant tasks. 
We propose to restructure the league into several tracks; each track will
focus on a different problem of real-world factory automation, such as 
scheduling and logistics, manipulation and assembly, or human-robot collaboration.
These tracks, once mature, will be merged to create a more complex and collaborative 
scenario, while allowing teams to focus on their specific area of expertise.

We begin in Section~\ref{sec:rcll} by summarizing the development and current state of the RCLL, after which 
we motivate the need for a reenvisioned industrial robotics competition in Section~\ref{sec:vision}. 
In Section~\ref{sec:tracks}, we introduce our initial proposal for new future tracks and which
tasks they might encompass, before we sketch in Section~\ref{sec:transition} a transition from the existing into the newly proposed competition, and then conclude in Section~\ref{sec:conclusion}. 

\section{The RoboCup Logistics League: The Story So Far}\label{sec:rcll}
The RoboCup Logistics League addresses a set of challenges for the development of autonomous robots
relevant to Industry 4.0 factory environments. In particular, the league focuses on a set of high-level
control tasks, like planning under uncertainty, job scheduling, multi-agent coordination, 
while also considering aspects of robust execution of skills on robots, and long-term autonomy. 
Robots need to be able to react in a flexible way to online requests, unexpected failures, and outages.
They furthermore need to be able to navigate a large-scale environment, use various mechanisms for
detecting and interacting with relevant machines and objects, avoid dynamic obstacles, and 
manipulate the products they're producing reliably. 

Since its inception in 2012, the RoboCup Logistics League (RCLL) has established itself
well in the Industrial track of the RoboCup, despite its high barrier of entry.
Over its history, the league managed to attract teams from 18 different universities across 9 countries.
The current roster consists of 7 active teams from Europe, North America, and Asia. 

Over the years, the competition has seen a significant evolution.
The original league used a ground-based puck transportation
scenario and separate fields for each team. Signal lights were used as stand-ins for machines, requiring only limited manipulation by the teams' robots.
Shared fields, introduced in 2014, increased the need
for precise localization and collision avoidance.
In 2015, the league 
introduced physical production machines based on Festo's Modular Production System stations that carried out
actual assembly tasks~\cite{niemuellerProposalAdvancementsLLSF2013}. This change also increased the complexity
of the planning, scheduling, and multi-agent coordination problems, by allowing for up to 240 product configurations. Furthermore, the need for robust manipulation added another
dimension of complexity to the competition. Following these changes in format, the following years were used
to increase the league's difficulty (e.g., putting more emphasis on timely orders, as well as the
introduction of competitive orders, markerless machine detection, and exploration of machines) based on the 
progress of teams. 

\begin{figure}
    \centering
    \includegraphics[width=1\linewidth]{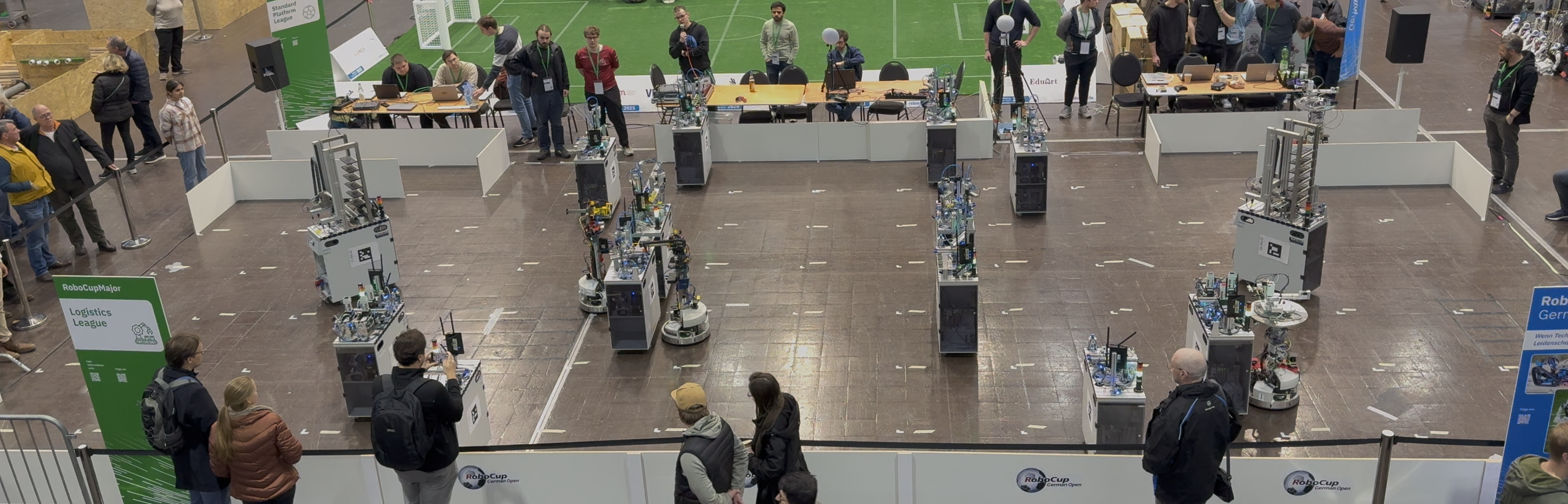}
    \caption{An RCLL game at the German Open 2025.
    }
    \label{fig:rcll-today}
\end{figure}

In the current setup, the league employs two parallel tracks to address the different development stages
of teams, and to reduce the barrier of entry for new teams. In the \textbf{challenge track}, teams can focus
on single tasks from the production scenario like manipulation, navigation, or machine recognition. 
Once these challenges have been solved, teams can qualify and progress towards the main track, where
the long-term production scenario is played.
In the \textbf{main track} (depicted in Figure \ref{fig:rcll-today}), in each game, two teams deploy a set of up to three robots playing against each other. 
Seven production machines per team are randomly placed on a shared field. 
Each of the machines represent a different production
step required to assemble the workpieces according to the order descriptions. 
During a 20 minute game, an identical set of 10 different orders with varying complexity is announced to the teams.
Scores are calculated based on the number of delivered products, their complexity (i.e., the number of machine 
interactions required), and the timeliness of their delivery.

To solve the league's challenges, multiple planning and coordination approaches have been developed since the inception of the competition, e.g., using a BDI architecture with HTNs~\cite{ulzRobustFlexibleSystem2019}, distributed as well as centralized goal reasoning ~\cite{niemuellerGoalReasoningCLIPS2019,hofmannWinningRoboCupLogistics2019,hofmannMultiagentGoalReasoning2021,swobodaUsingPromisesMultiagent2022,viehmannWinningRoboCupLogistics2023}, macro-action planning~\cite{debortoliImprovingApplicabilityPlanning2024}, temporal planning~\cite{schapersASPbasedTimeboundedPlanning2018,viehmannTransformingRoboticPlans2021,debortoliEnhancingTemporalPlanning2023,beikircherRobustIntegrationPlanning2024} and diagnosis for fault detection~\cite{haberingUsingPlatformModels2021,debortoliDiagnosisHiddenFaults2021}.
The planning problem incorporated in the RCLL also sparked interest in the planning community, leading to a planning competition at ICAPS~\cite{niemuellerPlanningCompetitionLogistics2016}. 
In addition to high-level reasoning, the competition also saw major improvements in other aspects such as manipulation with sophisticated gripper designs~\cite{beikircherRobustIntegrationPlanning2024,viehmannWinningRoboCupLogistics2023}, but also perception and low-level control~\cite{ulzRobustFlexibleSystem2019,tschescheUsingOfftheshelfDeep2025}.
Overall, these improvements lead to a significant increase in performance; while the top-scoring team in the year 2019 scored 94 points, the top score improved to 465 points in the year 2024.



\section{The Vision for RoboCup Smart Manufacturing}\label{sec:vision}

The participating teams in the RCLL have solved major challenges of industrial robotics, but the competition has so far focused on flexible manufacturing and lot-size-one production.
Recent research emphasizes the role of digital twins, AI-driven optimization, and human-centric design as critical enablers of smart manufacturing~\cite{sisinniIndustrialInternetThings2018,huIndustrialInternetThings2024,lengIndustry50Prospect2022,xuIndustry40Industry2021}.
Our vision builds on these pillars to create an evolving and modular competition, based on the following cornerstones:
\begin{description}
    \item[Cyber-physical integration.]
        The seamless integration of robots, machines, and sensors with digital systems is vital for real-time monitoring, decision-making, and optimization.
        While the RCLL has long been using an autonomous referee box~\cite{niemuellerProposalAdvancementsLLSF2013} and has since extended digital monitoring, e.g., by allowing teams to send their current status, the production status is only known roughly and the communication is not standardized.
    \item[Digital twins.]
        Continuous factory simulation is crucial for workflow optimization and analysis and is used for real-time prediction of process success, simulating alternate action plans, or fault detection.
        Although simulation has played a key role in the existing competition~\cite{zwillingSimulationRoboCupLogistics2014}, digital twins do not yet play a major role.
        We envision teams using a digital twin to monitor their robots in real-time, allowing for performance prediction 
        and monitoring.
    \item[Data-driven optimization.]
        Data analysis is an increasingly important topic in industry, where it can help streamline processes, remove redundancies, and increase effectiveness of operations \cite{dtaDrivenSmartManufacture} . 
        It can also help with improving production by creating better process time estimates and 
        detecting disruptions at an early stage. In the RCLL, relevant process data is already collected
        by the referee box. We want to encourage teams to use this information to improve their strategy. 
    \item[Intelligent decision making.]
        The production logistics in the current RCLL already focuses heavily on task planning and scheduling.
        As we envision production logistics to also be part of the new competition, intelligent decision making will remain a vital aspect of the competition.
    \item[Human-robot collaboration.]
        Especially with the shift towards \emph{Industry 5.0}, human-robot collaboration has become a central aspect of smart manufacturing~\cite{lengIndustry50Prospect2022}.
        Thus, instead of having a fully autonomous smart factory as the competition scenario, we propose a competition where robots need to cooperate with humans in order to manufacture complex products.
    \item[Interoperability and modularity.]
        A modular production system is essential in modern-day factories, as the whole system cannot be provided or maintained by a single supplier.
        Instead, robots, machines, and humans need to communicate via well-defined interfaces, e.g., OPC-UA or MQTT.
        We propose a modular competition, where each team can focus on a particular aspect, e.g., assembly, but the robots of different teams need to cooperate to run the overall factory.
        Adopting standard industry communication protocols allows for easier adoption of academic innovations in industrial setups and will also enhance industry participation.
\end{description}

To incorporate these cornerstones into the \emph{RoboCup Smart Manufacturing League}, we envision a competition with a modular, track-based structure.
Each track will focus on a particular aspect of a smart factory, e.g., transporation and product flow, product assembly,
and human-robot collaboration.
Each track targets a different type of robotic platform used in smart manufacturing, such as mobile robots, robot arms and humanoids. This way, each team may focus on a particular aspect and does not need to invest in multiple robot platforms.


Another key aspect will be standardized interfaces between the robots, allowing a seamless handover from one challenge to the next.
For example, a logistics robot needs to inform the assembly robot about a delivery, so it can continue with the next manufacturing step.
These task handovers may also include defect detection, naturally including additional aspects into the competition.


Initially, all the challenges will be independent to allow teams to get acquainted with the challenges, which will also simplify the change from the current competition format to the new smart manufacturing competition. Over time, as the performance of the participating teams increases, the challenges will be progressively integrated. Eventually, the competition will develop into a fully integrated smart factory, solving all major aspects of smart manufacturing.

\section{Tracks in RoboCup Smart Manufacturing}\label{sec:tracks}
We identified the following seven core robot skills for smart manufacturing environments:
\begin{description}
\item[Manipulation]
covers all interactions between the robot and objects. The SML will cover a broad set of different manipulation tasks that is representative of the large set of industrial applications. 
These tasks might include grasping different objects with variety in shape and material, picking from known and unknown places and placing objects at defined spots \cite{zhuDeformableObjects2022}.
Other constraints put on the robots could be immovable or movable obstacles and constrained workspaces requiring whole-body control.
\item[Low-level control] is required to execute the robot's movement and often requires a high computation frequency. A high precision in low-level control allows for more complex motions and therefore more complex tasks to execute \cite{zhuDeformableObjects2022,tschescheUsingOfftheshelfDeep2025}.
\item[Mobile navigation] includes all navigation aspects required to move with one or multiple mobile robots like localization, dynamic obstacle avoidance, and multi-agent pathfinding. In industry settings, planning and executing conflict-free routes for multiple robots is essential~\cite{macenskiROS2Navigation2023,borseRobotinoNavigation2024}.
\item[Task planning and Scheduling] refers to a system's capability to plan multiple actions into the future to maximize its reward. The necessity for this skill depends on the possible options within the tracks. In tracks with low focus on this skill, there might be only one goal and no options to choose from. At the same time, the production logistics track
requires to make complex reasoning to select among the available actions 
~\cite{hofmannMultiagentGoalReasoning2021,beikircherRobustIntegrationPlanning2024}.
\item[Perception] 
is a skill required for nearly all robotics tasks but nonetheless differs in its complexity between this league's tracks. Visual sensors such as cameras are used to identify objects or areas and localize them in the image as well as in the real world to enable interactions with them~\cite{chenHumanInTheLoop2025,tschescheUsingOfftheshelfDeep2025}.
\item[Human-robot collaboration and safety] is an increasingly important field in industry with the integration of cobots in shared spaces with humans. 
This introduces a lot of challenges to the robot not only to
ensure safety when working alongside humans,
but also to collaborate with them in a natural way. Therefore, the focus should be on verbal communication and gesture \cite{chenHumanInTheLoop2025,mathesonHRC2019}.
\item[Agility] describes the ability of robots to perform complex motion like moving through rough terrain, over difficult obstacles \cite{mohammad2024}, or to perform granular manipulation. Many conventional factory environments are still designed for humans, and thus don't allow for easy integration of robots into their workflows.
For example, an older factory might make excessive use of stairs and ladders, or has complex control panels that require precise manipulations.
\end{description}

\begin{figure}
  \centerline{\includegraphics[page=4]{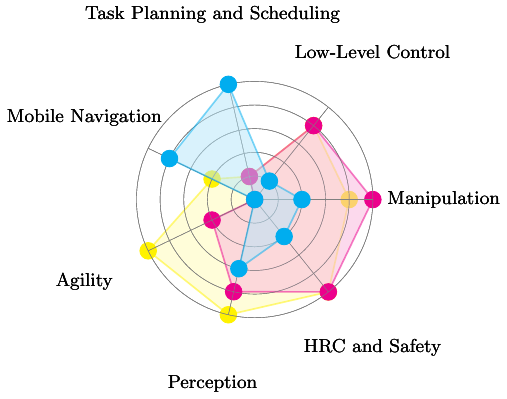}}
    \caption{Skill characterizations for each track. Cyan for production logistics, magenta for assembly, and yellow for the humanoid track. HRC stands for Human-Robot Collaboration.}
    \label{fig:skills}
\end{figure}

Our goal is to address the full spectrum of core skills in RoboCup SML through three distinct tracks: production logistics, assembly, and humanoid manufacturing. Each track emphasizes a different subset of these skills, as illustrated in Figure~\ref{fig:skills}.
All tracks will share a common base framework, such as an equal game duration, standardized communication interfaces and a common arena (although initially in separate areas). This enables the gradual integration of the individual tracks into a unified production scenario in the future. 

\subsection{Production Logistics}
Planning and scheduling of production steps, as well as coordination of material flow remains an important
problem in smart factory environments. The RoboCup SML's Production Logistics track naturally emerges from
the current main track of the RCLL. The focus of this track will remain on long-term autonomy, robot-robot collaboration and fleet coordination, as well as the core planning and scheduling of online orders. 

However, we further want to evolve this part of the league by requiring increasing flexibility and variety
in the workflow. By replacing the current Festo machines with more cost effective solutions, we can broaden the types of different stations available to the robots allowing
for more customized production flows, where more dynamic decisions are required. This may include more disruptive events, such as faulty assembly steps, requests from customers, and dynamically evolving shopfloors, e.g., paths being blocked or opened up over time.

Similarly, the new arena will impose new challenges when navigating the shopfloor. Dynamic obstacles may need to be classified to gauge required safety distances and coordination with robots of unknown shapes might be required to avoid bottlenecks in tight corridors in between the different production areas.
Furthermore, teams may also deploy a heterogeneous fleet of mobile robots, e.g., having a robot that can store multiple products and can hand over material to other robots to speed up production by optimizing task parallelization.

\subsection{Assembly}
The Assembly track focuses on manipulation tasks and will require 6DoF robot arms. These robots need to assemble multiple workpieces starting with the current RCLL workpieces. In the near future, these track's robots should be able to fulfil the current work of the logistics track's Modular Production
System stations and replace them.
This includes detecting and localizing workpieces, rings, and caps through a camera as well controlling the arm to assemble the requested results.
These non-reflective object have a known cylindrical shape. Therefore they are easy to detect and manipulate. 

After archiving this goal, the next goal will be to assemble more complex products and focus more on shared assembly between humans and robots.
The products can have more complex shapes that can only be grasped at different positions and might need to be placed differently before grasping. The object surface can be more reflective or blend more into the background like wood products on a table.
The track could also include soft or slippery objects and objects unknown to the teams. 
The problem this track tries to solve is general manipulation and therefore the uncertainty in the targetted objects should increase over time.
The robot might need to change its end-effector for object classes like changing to a soft gripper for soft objects \cite{zhuDeformableObjects2022}.
Since the aim of this track is the assembly of many complex products without separating the track into multiple sub-tracks, the track will start with easy objects to pick up and place into a bin and increase continuously in difficulty such that fast manipulation is favoured.

Another important aspect in this track is human-robot collaboration which will be integrated by sharing a workspace between the robot arm and humans, splitting up work on the same object, commanding the robot by voice, and working together to fulfill a task. The track is closely related to the tasks found in the existing @Work league, and will be developed in cooperation with them.

\subsection{Humanoid Manufacturing}
Humanoid robots are the most natural choice for gradually replacing human workers by robots and have seen great interest in industry~(e.g.,~\cite{chesnokovaFutureWork52024,heaterMercedesBeginsPiloting2024,saeedy$40BillionStartup2025}).
%
A core use-case for humanoids in manufacturing scenarios is in shop floors that are centered around human usage, where robots with simpler kinematics are not suitable for deployment.
Tasks in that area may include the operation of a warehouse (packaging goods and delivering them outside of the shop floor), assembly assistance (moving mobile tool containers, opening drawers, replenishing material), as well as operating in traditional factories that deploy production machines with human interfaces (buttons, touch screens, levers, staircases).

Another area currently dominated by humans is quality control and monitoring:
Supervising the shop floor, detecting issues with assembly machines or products and fixing or reporting them.
This may also require calling in human support in situations that requires a human decision, hence human robot collaboration will also play a big role.

Lastly, as humanoid robots should be able to perform the physical labor of a human worker, the ability to learn from human supervisors when working on new tasks is a core skill to leverage the agility of humanoid platforms in a broad range of use-cases.



\section{Transitioning from the RCLL to the SML}\label{sec:transition}
By extending the range of tasks compared to the RCLL, some consideration is required in order to ensure a smooth transition towards the proposed tracks.

Firstly, compared to the RCLL, the SML needs a more heterogenous arena, where different tracks can be co-located. Short-term, this should allow tracks to work independently from each other, until meaningful collaboration interfaces can be deployed (based on the capabilities of participating teams).
In particular, the factory setup of the RCLL based on Festo Didactics Modular Production System needs to be replaced by a more easily extensible and cost-effective solution, where the deployment of the arena can be feasibly handled by the local organizers and the participating teams without too much additional overhead.
The idea is that stations are kept simple in their minimum requirements, e.g., the base station of the RCLL may be replaced by an ordinary table, where parts of different properties are simply placed at and robots may directly take them.
At the same time, they should also enable collaboration between different tracks in the long term, e.g., a task in the assembly track could be to sort and place different objects at the new base station table to be transported.

A second important step to establish the tracks is to define platform regulations and requirements that allow teams to invest in robot platforms suitable for the current and future tasks of the tracks.
Example requirements could be a minimal required actuation reach or defined interaction points. These requirements may be revised in the future as the league continues to develop, but should cover the current and future tasks that may arise, such that teams can develop their robotic platforms accordingly.
The Committees of the RCLL are also coordinating with the @Work league while shaping the tracks, as both leagues see the potential to join efforts when building up the SML. The @Work league is already emphasizing manipulation tasks in industrial use-cases, hence it seems like a natural fit to shape the manipulation track through the ideas of @Work, similar to the RCLL becoming the starting point of the production logistics track. In this way, the RoboCup Industrial leagues could unify in the SML and align to a common vision.
To enable a frictionless integration and reduce the burden on teams, the respective tracks should build heavily on the existing prior leagues and maintain full robot hardware compatibility in their first iteration.
Subsequent evolution of the tracks may lead to updated hardware requirements, but should always consider the currently used platforms and provide a feasible upgrade path with sufficient prior notice. 

Teams will be able to freely choose in which of the tracks to compete and we expect existing teams, especially in the transition phase, to only compete in the track related to their prior league. Some version of a qualification procedure for each track will be required, where teams need to prove that their robots are capable of performing entry-level tasks of the track.





\section{Conclusion}\label{sec:conclusion}
In this paper, we presented our vision for a new and comprehensive RoboCup Industrial 
competition, the \emph{RoboCup Smart Manufacturing League (SML)}. This new league aims to address a much wider
spectrum of relevant industrial challenges, including (humanoid) manipulation, human-robot collaboration, adaptive manufacturing, but also the existing long-term production scenario. 

We identified seven core robot skills that are required in modern and future production facilities. Our new competition format will address these with three tracks, each focusing on different core skills. 
These new tracks serve as incubators for new teams with different specialisations, enabling focused research and development on the individual competencies. However, once matured, our long term-goal is to fully integrate the tracks into a comprehensive competition, in which teams with different skills need to actively collaborate. 

With the SML, we aim not only to lower the barrier to entry for new participants by allowing them to start in a focused track, but also to attract teams from a wide range of research backgrounds including AI, robotics, automation, and human-robot collaboration. The eventual integration of the tracks will encourage cross-disciplinary teamwork, a significant evolution in the RoboCup competition structure.

\begin{credits}
\subsubsection{\ackname}
The research was funded by the Alexander von Humboldt Foundation with funds from Federal Ministry of Research, Technology and Space (BMFTR) and the Ministry of Culture and Science of the German State of North Rhine-Westphalia (MKW) under the Excellence Strategy of the Federal Government and the L\"ander, the Deutsche Forschungsgemeinschaft (DFG, German Research Foundation) under Germany's Excellence Strategy – EXC-2023 Internet of
Production – 390621612, the EU ICT-48 2020 project TAILOR (No.\  952215), Research Training Group 2236 (UnRAVeL), and BMFTR under grant no 02L19C602 and 02L19C400. We thank Department 5 of FH Aachen for their support.
\end{credits}

\bibliography{till, rcsm}

\begin{thebibliography}{10}
\providecommand{\url}[1]{\texttt{#1}}
\providecommand{\urlprefix}{URL }
\providecommand{\doi}[1]{https://doi.org/#1}

\bibitem{beikircherRobustIntegrationPlanning2024}
Beikircher, D., Bortoli, M.D., F{\"u}rba{\ss}, L., Kernbauer, T., Kohout, P.,
  Lampel, D., Masiero, A., Moser, S., Nagele, M., {Steinbauer-Wagner}, G.:
  Robust {{Integration}} of~{{Planning}}, {{Execution}}, {{Recovery}}
  and~{{Testing}} to~{{Win}} the~{{RoboCup Logistics League}}. In: {{RoboCup}}
  2023: {{Robot World Cup XXVI}}. pp. 362--373. Springer Nature Switzerland,
  Cham (2024). \doi{10.1007/978-3-031-55015-7_30}

\bibitem{borseRobotinoNavigation2024}
Borse, S., Viehmann, T., Ferrein, A., Lakemeyer, G.: A ros 2-based navigation
  and simulation stack for the robotino. In: Barros, E., Hanna, J.P., Okada,
  H., Torta, E. (eds.) RoboCup 2024: Robot World Cup XXVII. pp. 19--31.
  Springer Nature Switzerland, Cham (2025)

\bibitem{chenHumanInTheLoop2025}
Chen, H., Li, S., Fan, J., Duan, A., Yang, C., Navarro-Alarcon, D., Zheng, P.:
  Human-in-the-loop robot learning for smart manufacturing: A human-centric
  perspective. IEEE Transactions on Automation Science and Engineering
  \textbf{22},  11062--11086 (2025). \doi{10.1109/TASE.2025.3528051}

\bibitem{chesnokovaFutureWork52024}
Chesnokova, S.: Future of work: 5 things to know about {{Agility Robotics}}'
  {{Digit}}, a humanoid autonomous robot disrupting warehouses --- {{TFN}}.
  Tech Funding News  (Dec 2024),
  \url{https://techfundingnews.com/future-of-work-5-things-to-know-about-agility-robotics-digit-a-humanoid-autonomous-robot-disrupting-warehouses/}

\bibitem{debortoliEnhancingTemporalPlanning2023}
De~Bortoli, M., Chrpa, L., Gebser, M., {Steinbauer-Wagner}, G.: Enhancing
  {{Temporal Planning}} by~{{Sequential Macro-Actions}}. In: Logics in
  {{Artificial Intelligence}}. pp. 595--604. Springer Nature Switzerland, Cham
  (2023). \doi{10.1007/978-3-031-43619-2_40}

\bibitem{debortoliImprovingApplicabilityPlanning2024}
De~Bortoli, M., Chrpa, L., Gebser, M., {Steinbauer-Wagner}, G.: Improving
  {{Applicability}} of~{{Planning}} in~the~{{RoboCup Logistics League Using
  Macro-actions Refinement}}. In: {{RoboCup}} 2023: {{Robot World Cup XXVI}}.
  pp. 287--298. Springer Nature Switzerland, Cham (2024).
  \doi{10.1007/978-3-031-55015-7_24}

\bibitem{debortoliDiagnosisHiddenFaults2021}
De~Bortoli, M., Munoz~Gutierrez, S., {Steinbauer-Wagner}, G.: Diagnosis of
  hidden faults in the {{RCLL}}: 32nd {{International Workshop}} on
  {{Principle}} of {{Diagnosis}}. In: 32nd {{International Workshop}} on
  {{Principle}} of {{Diagnosis}} (Sep 2021)

\bibitem{haberingUsingPlatformModels2021}
Habering, D., Hofmann, T., Lakemeyer, G.: Using platform models for a guided
  explanatory diagnosis generation for mobile robots. In: Proceedings of the
  30th {{International Joint Conference}} on {{Artificial Intelligence}}
  ({{IJCAI}}). pp. 1908--1914 (2021). \doi{10.24963/ijcai.2021/263}

\bibitem{heaterMercedesBeginsPiloting2024}
Heater, B.: Mercedes begins piloting {{Apptronik}} humanoid robots. TechCrunch
  (Mar 2024),
  \url{https://techcrunch.com/2024/03/15/mercedes-begins-piloting-apptronik-humanoid-robots/}

\bibitem{hofmannWinningRoboCupLogistics2019}
Hofmann, T., Limpert, N., Matar{\'e}, V., Ferrein, A., Lakemeyer, G.: Winning
  the {{RoboCup Logistics League}} with fast navigation, precise manipulation,
  and robust goal reasoning. In: {{RoboCup}} 2019: {{Robot World Cup XXIII}}.
  pp. 504--516. Springer (2019). \doi{10.1007/978-3-030-35699-6_41}

\bibitem{hofmannMultiagentGoalReasoning2021}
Hofmann, T., Viehmann, T., Gomaa, M., Habering, D., Niemueller, T., Lakemeyer,
  G.: Multi-agent goal reasoning with the {{CLIPS Executive}} in the {{RoboCup
  Logistics League}}. In: Proceedings of the 13th {{International Conference}}
  on {{Agents}} and {{Artifical Intelligence}} ({{ICAART}}). vol.~1, pp.
  80--91. SciTePress (2021). \doi{10.5220/0010252600800091}

\bibitem{huIndustrialInternetThings2024}
Hu, Y., Jia, Q., Yao, Y., Lee, Y., Lee, M., Wang, C., Zhou, X., Xie, R., Yu,
  F.R.: Industrial {{Internet}} of {{Things Intelligence Empowering Smart
  Manufacturing}}: {{A Literature Review}}. IEEE Internet of Things Journal
  \textbf{11}(11),  19143--19167 (Jun 2024). \doi{10.1109/JIOT.2024.3367692}

\bibitem{lengIndustry50Prospect2022}
Leng, J., Sha, W., Wang, B., Zheng, P., Zhuang, C., Liu, Q., Wuest, T.,
  Mourtzis, D., Wang, L.: Industry 5.0: {{Prospect}} and retrospect. Journal of
  Manufacturing Systems  \textbf{65},  279--295 (Oct 2022).
  \doi{10.1016/j.jmsy.2022.09.017}

\bibitem{dtaDrivenSmartManufacture}
Li, W., Liang, Y., Wang, S.: Data Driven Smart Manufacturing Technologies and
  Applications (01 2021). \doi{10.1007/978-3-030-66849-5}

\bibitem{macenskiROS2Navigation2023}
Macenski, S., Moore, T., Lu, D., Merzlyakov, A., Ferguson, M.: From the desks
  of ros maintainers: A survey of modern \& capable mobile robotics algorithms
  in the robot operating system 2. Robotics and Autonomous Systems  (2023)

\bibitem{mathesonHRC2019}
Matheson, E., Minto, R., Zampieri, E.G.G., Faccio, M., Rosati, G.:
  Human–robot collaboration in manufacturing applications: A review. Robotics
   \textbf{8}(4) (2019). \doi{10.3390/robotics8040100},
  \url{https://www.mdpi.com/2218-6581/8/4/100}

\bibitem{mohammad2024}
Mohammad, N., Higgins, J., Bezzo, N.: A gp-based robust motion planning
  framework for agile autonomous robot navigation and recovery in unknown
  environments. In: 2024 IEEE International Conference on Robotics and
  Automation (ICRA). pp. 2418--2424 (2024).
  \doi{10.1109/ICRA57147.2024.10610382}

\bibitem{niemuellerGoalReasoningCLIPS2019}
Niemueller, T., Hofmann, T., Lakemeyer, G.: Goal reasoning in the {{CLIPS
  Executive}} for integrated planning and execution. In: Proceedings of the
  29th {{International Conference}} on {{Automated Planning}} and
  {{Scheduling}} ({{ICAPS}}). pp. 754--763 (2019),
  \url{https://kbsg.rwth-aachen.de/~hofmann/papers/clips-exec-icaps19.pdf}

\bibitem{niemuellerPlanningCompetitionLogistics2016}
Niemueller, T., Karpas, E., Vaquero, T., Timmons, E.: Planning competition for
  logistics robots in simulation. In: 4th {{ICAPS Workshop}} on {{Planning}}
  and {{Robotics}} ({{PlanRob}}) (2016)

\bibitem{niemuellerProposalAdvancementsLLSF2013}
Niemueller, T., Lakemeyer, G., Ferrein, A., Reuter, S., Ewert, D., Jeschke, S.,
  Penksy, D., Karras, U.: Proposal for advancements to the {{LLSF}} in 2014 and
  beyond. In: 1st {{Workshop}} on {{Developments}} in {{RoboCup Leagues}}
  (2013)

\bibitem{saeedy$40BillionStartup2025}
Saeedy, Berber Jin {and}~Alexander, E.G.: The \$40 {{Billion Startup Mystery
  Shaking Up Silicon Valley}}. WSJ  (Apr 2025),
  \url{https://www.wsj.com/tech/the-hottest-pre-ipo-stock-an-ai-robotics-startup-with-bold-claims-little-revenue-b0c1f03b}

\bibitem{schapersASPbasedTimeboundedPlanning2018}
Sch{\"a}pers, B., Niemueller, T., Lakemeyer, G., Gebser, M., Schaub, T.:
  {{ASP-based}} time-bounded planning for logistics robots. In: Proceedings of
  the 28th {{International Conference}} on {{Automated Planning}} and
  {{Scheduling}} ({{ICAPS}}). pp. 509--517. AAAI Press (2018),
  \url{https://aaai.org/ocs/index.php/ICAPS/ICAPS18/paper/view/17777}

\bibitem{sisinniIndustrialInternetThings2018}
Sisinni, E., Saifullah, A., Han, S., Jennehag, U., Gidlund, M.: Industrial
  {{Internet}} of {{Things}}: {{Challenges}}, {{Opportunities}}, and
  {{Directions}}. IEEE Transactions on Industrial Informatics  \textbf{14}(11),
   4724--4734 (Nov 2018). \doi{10.1109/TII.2018.2852491}

\bibitem{swobodaUsingPromisesMultiagent2022}
Swoboda, D., Hofmann, T., Viehmann, T., Lakemeyer, G.: Towards using promises
  for multi-agent cooperation in goal reasoning. In: {{ICAPS Workshop}} on
  {{Planning}} and {{Robotics}} ({{PlanRob}}) (2022).
  \doi{10.48550/arXiv.2206.09864}

\bibitem{tschescheUsingOfftheshelfDeep2025}
Tschesche, M., Hofmann, T., Ferrein, A., Lakemeyer, G.: Using off-the-shelf
  deep neural networks for~position-based visual servoing. In: {{RoboCup}}
  2024: {{Robot World Cup XXVII}}. pp. 32--43. Springer Nature Switzerland,
  Cham (2025). \doi{10.1007/978-3-031-85859-8_3}

\bibitem{ulzRobustFlexibleSystem2019}
Ulz, T., Ludwiger, J., Steinbauer, G.: A {{Robust}} and {{Flexible System
  Architecture}} for {{Facing}} the {{RoboCup Logistics League Challenge}}. In:
  {{RoboCup}} 2018: {{Robot World Cup XXII}}. pp. 488--499. Springer
  International Publishing, Cham (2019). \doi{10.1007/978-3-030-27544-0_40}

\bibitem{viehmannTransformingRoboticPlans2021}
Viehmann, T., Hofmann, T., Lakemeyer, G.: Transforming robotic plans with timed
  automata to solve temporal platform constraints. In: Proceedings of the 30th
  {{International Joint Conference}} on {{Artificial Intelligence}}
  ({{IJCAI}}). pp. 2083--2089 (2021). \doi{10.24963/ijcai.2021/287}

\bibitem{viehmannWinningRoboCupLogistics2023}
Viehmann, T., Limpert, N., Hofmann, T., Henning, M., Ferrein, A., Lakemeyer,
  G.: Winning the~{{RoboCup Logistics League}} with~{{Visual Servoing}}
  and~{{Centralized Goal Reasoning}}. In: {{RoboCup}} 2022: {{Robot World Cup
  XXV}}. pp. 300--312. Springer International Publishing, Cham (2023).
  \doi{10.1007/978-3-031-28469-4_25}

\bibitem{xuIndustry40Industry2021}
Xu, X., Lu, Y., {Vogel-Heuser}, B., Wang, L.: Industry 4.0 and {{Industry}}
  5.0---{{Inception}}, conception and perception. Journal of Manufacturing
  Systems  \textbf{61},  530--535 (Oct 2021). \doi{10.1016/j.jmsy.2021.10.006}

\bibitem{zhuDeformableObjects2022}
Zhu, J., Cherubini, A., Dune, C., Navarro-Alarcon, D., Alambeigi, F., Berenson,
  D., Ficuciello, F., Harada, K., Kober, J., Li, X., Pan, J., Yuan, W.,
  Gienger, M.: Challenges and outlook in robotic manipulation of deformable
  objects. IEEE Robotics \& Automation Magazine  \textbf{29}(3),  67--77
  (2022). \doi{10.1109/MRA.2022.3147415}

\bibitem{zwillingSimulationRoboCupLogistics2014}
Zwilling, F., Niemueller, T., Lakemeyer, G.: Simulation for the {{RoboCup
  Logistics League}} with {{Real-World Environment Agency}} and {{Multi-level
  Abstraction}}. In: {{RoboCup Symposium}}. Jo{\~a}o Pessoa, Brazil (2014)

\end{thebibliography}
\bibliographystyle{splncs04}

\end{document}